%% file: acl_latex.tex
\pdfoutput=1
\documentclass[11pt]{article}

\usepackage[]{acl}

\usepackage{times}
\usepackage{latexsym}

\usepackage[T1]{fontenc}
\usepackage[utf8]{inputenc}

\usepackage{microtype}

\usepackage{amsmath,amssymb}
\usepackage{booktabs}
\usepackage{comment}
\usepackage{graphicx}
\usepackage{multirow}

\input macros.tex

\title{Question Answering Infused Pre-training\\ of General-Purpose Contextualized Representations}

\author{
Robin Jia\thanks{\quad Work done while a visiting researcher at Facebook AI Research.} \\
University of Southern California \\
\texttt{robinjia@usc.edu} \\\And
Mike Lewis, Luke Zettlemoyer \\
Facebook AI Research \\
\texttt{\{mikelewis,lsz\}@fb.com}
}

\begin{document}
\maketitle
\begin{abstract}
\input abstract.tex
\end{abstract}

\input introduction.tex
\input approach.tex
\input experiments.tex
\input discussion.tex

\section*{Acknowledgements}
We thank Terra Blevins for investigating applications to word sense disambiguation,
Jiaxin Huang for providing the few-shot NER splits used in their paper,
and Douwe Kiela, Max Bartolo, Sebastian Riedel, Sewon Min, Patrick Lewis, Scott Yih, and our anonymous reviewers for their feedback.

% Entries for the entire Anthology, followed by custom entries
\bibliography{anthology,custom}
\bibliographystyle{acl_natbib}

\appendix
\input appendix.tex

\end{document}

%% file: macros.tex
\newcommand{\nl}[1]{``\textit{#1}''}
\newcommand{\quip}{\textsc{QuIP}}
\newcommand{\NoQgen}{Bi-encoder + MRQA}
\newcommand{\Teacher}{Cross-encoder + MRQA}
\newcommand{\NoTeacher}{\quip, no teacher}
\newcommand{\CrossStudent}{\quip, cross-encoder student}
\newcommand{\Unsup}{Bi-encoder + UnsupervisedQA}

%% file: abstract.tex
We propose a pre-training objective based on question answering (QA) for learning general-purpose contextual representations,
motivated by the intuition that the representation of a phrase in a passage should encode all questions that the phrase can answer in context.
To this end, we train a bi-encoder QA model, which independently encodes passages and questions, to match the predictions of a more accurate cross-encoder model on 80 million synthesized QA pairs.
By encoding QA-relevant information, the bi-encoder's token-level representations are useful for non-QA downstream tasks without extensive (or in some cases, any) fine-tuning.
We show large improvements over both RoBERTa-large and previous state-of-the-art results on zero-shot and few-shot paraphrase detection on four datasets, few-shot named entity recognition on two datasets, and zero-shot sentiment analysis on three datasets.

%% file: introduction.tex
\section{Introduction}
While masked language models build contextualized word representations, they are pre-trained with losses that minimize distance to \emph{uncontextualized} word embeddings \citep{peters-etal-2018-deep,devlin-etal-2019-bert,liu2019roberta}.
This objective yields a good initialization for downstream fine-tuning, but the pre-trained representations themselves are not optimized for being immediately useful without fine-tuning.
In this paper, we introduce Question Answering Infused Pre-training (\quip),
a new pre-training loss based on question answering (QA) that depends much more directly on context. 
\quip{} learns improved token-level representations that are useful in zero-shot and few-shot settings, where extensive fine-tuning is not possible.

\begin{figure}
\begin{center}
\includegraphics[width=\linewidth]{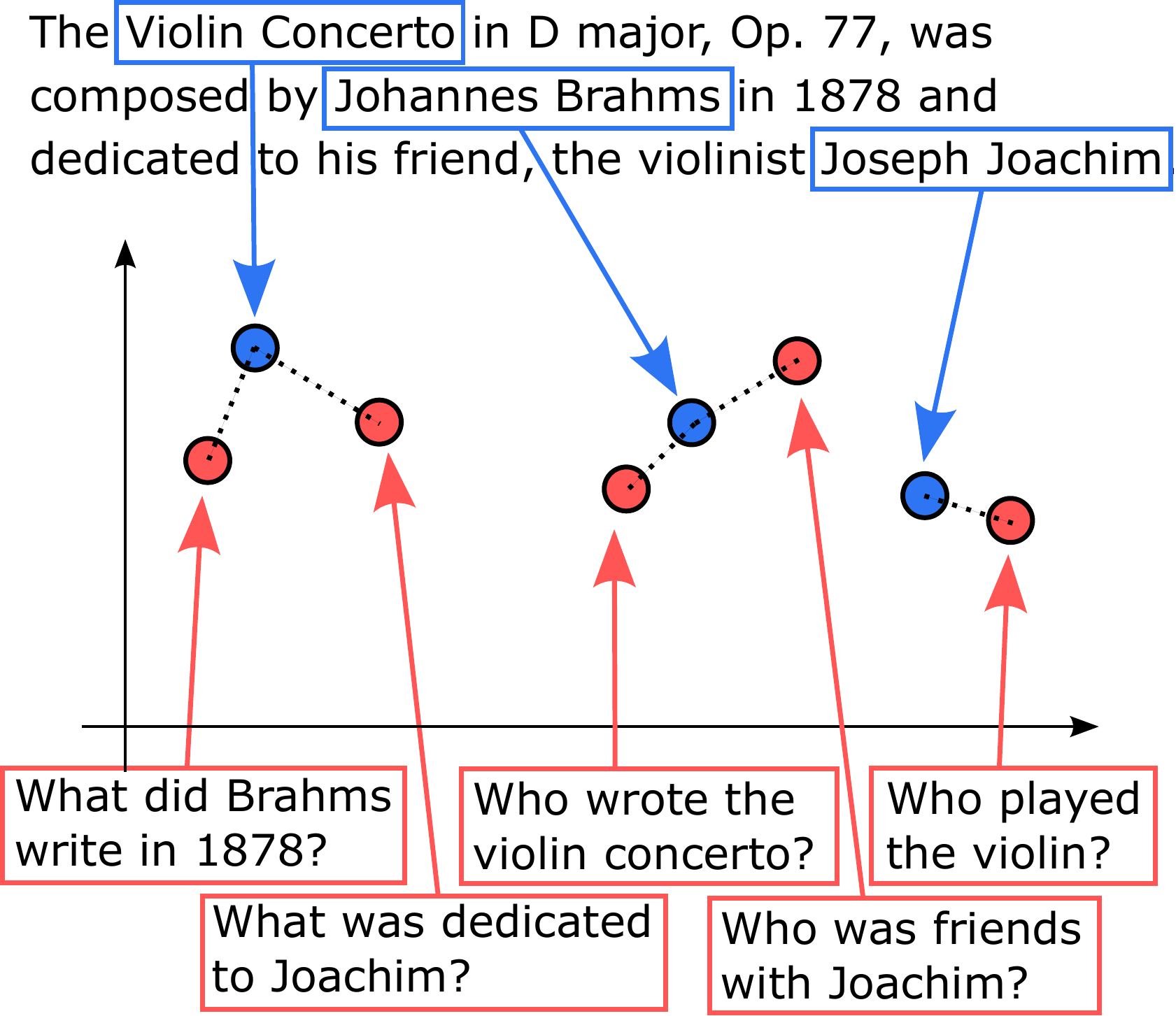}
\end{center}
\caption{An overview of Question Answering Infused Pre-training.
Our model independently creates vector representations (middle) for phrases in a passage (top) and for synthesized questions (bottom).
Our objective encourages the vector for each phrase to have high similarity with the vectors for all questions it answers.
}
\label{fig:overview}
\end{figure}

Our intuition for \quip{} is that the contextualized representation for a phrase in a passage should contain enough information to identify all the questions that the phrase could answer in context. For example, in Figure \ref{fig:overview}, the representation for \emph{Johannes Brahms} should be similar to the representation of all questions it can answer, such as \nl{Who wrote the violin concerto?}
We anticipate that optimizing passage representations for QA should benefit many downstream tasks,
as question-answer pairs have been used as broad-coverage meaning representations \citep{he-etal-2015-question,michael-etal-2018-crowdsourcing}, and a wide range of NLP tasks can be cast as QA problems \citep{levy-etal-2017-zero,mccann2018natural,gardner2019question}.
For instance, our learned representations should encode whether a phrase answers a question like \nl{Why was the movie considered good?}, which corresponds to identifying rationales for sentiment analysis.

We train \quip{} with a bi-encoder extractive QA objective.
The model independently encodes passages and questions such that the representation of each phrase in a passage is similar to the representation of reading comprehension questions answered by that phrase.
We use a question generation model to synthesize 80 million QA examples, then train the bi-encoder to match the predictions of a cross-encoder QA model, which processes the passage and question together, on these examples.

Bi-encoder QA has been used before for efficient open-domain QA via phrase retrieval \citep{seo-etal-2018-phrase,seo-etal-2019-real,lee-etal-2020-contextualized,lee2021learning},
but its lower accuracy compared to cross-encoder QA has previously been viewed as a drawback.
We instead view the relative weakness of bi-encoder QA as an opportunity to improve contextual representations via knowledge distillation, as self-training can be effective when the student model must solve a harder problem than the teacher \citep{xie2020self}.
In particular, since the bi-encoder does not know the question when encoding the passage, it must produce a single passage representation that simultaneously encodes the answers to \emph{all} possible questions.
In contrast, while cross-encoder QA models are more accurate, they depend on a specific question when encoding a passage; thus, they are less suited to downstream use cases that require contextualized representations of passages in isolation.

We show that \quip{} token-level representations are useful in a variety of zero-shot and few-shot learning settings, both because the representations directly encode useful contextual information, and because we can often reduce downstream tasks to QA.
For few-shot paraphrase detection, \quip{} with BERTScore-based features \citep{zhang2020bertscore} outperforms prior work by $9$ F1 points across four datasets.
For few-shot named entity recognition (NER),
\quip{} combined with an initialization scheme that uses question embeddings improves over RoBERTa-large by $14$ F1 across two datasets.
Finally, for zero-shot sentiment analysis, \quip{} with question prompts improves over RoBERTa-large with MLM-style prompts by $5$ accuracy points across three datasets, and extracts interpretable rationales as a side effect.
Through ablations, we show that using real questions, a strong teacher model, and the bi-encoder architecture are all crucial to the success of \quip.
Other design decisions (e.g., question generation decoding strategies) do not qualitatively affect our main findings, pointing to the stability of the \quip{} approach.\footnote{Code to reproduce all results can be found at \url{https://github.com/facebookresearch/quip}.}

%% file: approach.tex
\section{QA Infused Pre-training}
\label{sec:pretraining}
QA Infused Pre-training (\quip) involves pre-training contextual representations with a bi-encoder extractive QA objective.
In contrast with masked language modeling, \quip{}'s training objective directly encourages contextual representations to encode useful semantic information, namely information about what questions can be answered by each span.
In contrast with a cross-encoder QA model, \quip{}'s bi-encoder is trained to encode single passages rather than passage-question pairs, making it more transferable to tasks involving single passages.
Moreover, \quip{} learns to push each phrase's representation far away from those of questions the phrase does \emph{not} answer;
this ability to represent unanswerability is crucial for correctly handling some question-based prompts.

We now introduce some basic notation (\S\ref{sec:notation}),
then describe the \quip{} pipeline, which consists of three steps:
question generation (\S\ref{sec:generation}), cross-encoder teacher re-labeling (\S\ref{sec:teacher}), and bi-encoder training (\S\ref{sec:biencoder}).

\subsection{Notation}
\label{sec:notation}
All models operate on sequences of tokens $x = [x_1, \dotsc, x_L]$ of length $L$. By convention, we assume that $x_1$ is always the special beginning-of-sequence token.
We learn an encoder $r$ that maps inputs $x$ to outputs $r(x) = [r(x)_1, \dotsc, r(x)_L]$ where each $r(x)_i \in \mathbb{R}^d$ for some fixed dimension $d$.
We call $r(x)_i$ the contextual representation of the $i$-th token in $x$.

In extractive question answering, a model is given a context passage $c$ and question $q$, and must output a span of $c$ that answers the question. Typically, models independently predict probability distributions $p(a_{\text{start}} \mid c, q)$ and $p(a_{\text{end}} \mid c, q)$
over the answer start index $a_{\text{start}}$ and end index $a_{\text{end}}$.

\subsection{Question Generation}
\label{sec:generation}
\paragraph{Question generation model.} We train a BART-large model \citep{lewis-etal-2020-bart} to generate question-answer pairs given context passages. The model receives the passage as context and must generate the answer text, then a special separator token, then the question. This approach is simpler than prior approaches that use separate models for answer and question generation \citep{lewis2019generative,alberti-etal-2019-synthetic,puri-etal-2020-training}, and works well in practice.

\paragraph{Training data.} We train on the training data from the MRQA 2019 Shared Task \citep{fisch-etal-2019-mrqa}, which includes six datasets: HotpotQA \citep{yang-etal-2018-hotpotqa}, NaturalQuestions \citep{kwiatkowski-etal-2019-natural}, NewsQA \citep{trischler-etal-2017-newsqa}, SearchQA \citep{dunn2017searchqa}, SQuAD \citep{rajpurkar-etal-2016-squad}, and TriviaQA \citep{joshi-etal-2017-triviaqa}.
These datasets cover many of the text sources commonly used for pre-training \citep{liu2019roberta,lewis-etal-2020-bart}, namely Wikipedia (HotpotQA, NaturalQuestions, SQuAD), News articles (NewsQA), and general web text (SearchQA, TriviaQA).

\paragraph{Generating questions.} We run our question generation model over a large set of passages to generate a large dataset of question-answer pairs. We decode using nucleus sampling \citep{holtzman2020curious} with $p=0.6$, which was chosen by manual inspection to balance diversity with quality of generated questions.
We do not filter questions in any way.
While we observed some flaws related to question quality (questions were not always well-formed) and diversity (for some passages, the same or very similar questions were asked multiple times), this approach nonetheless yielded good downstream results.
Attempts to mitigate these issues, such as using a two-stage beam search to ensure that questions for the same passage have different answers, did not noticeably change our downstream results (see \S\ref{sec:stability}).
We obtain passages from the same training corpus as RoBERTa \citep{liu2019roberta}, which uses four sub-domains: \textsc{BookCorpus} plus Wikipedia, \textsc{CC-News}, \textsc{OpenWebText}, and \textsc{Stories}.
For each domain, we sample 2 million passages and generate 10 questions per passage, for a total of 80 million questions.\footnote{We estimate that using the entire corpus with these settings would generate around 900 million questions. We leave investigation of further scaling to future work.}

\subsection{Teacher Re-labeling}
\label{sec:teacher}
The answers generated by our BART model are not always accurate, nor are they always spans in the context passage.
To improve the training signal, we re-label examples with a teacher model, as is common in knowledge distillation \citep{hinton2015distilling}.
We use a standard cross-encoder RoBERTa-large model trained on the MRQA training data as our teacher.
The model takes in the concatenation of the context passage $c$ and question $q$ and predicts $a_{\text{start}}$ and $a_{\text{end}}$ with two independent 2-layer multi-layer perceptron (MLP) heads.
We denote the teacher's predicted probability distribution over $a_{\text{start}}$ and $a_{\text{end}}$ as
$T_{\text{start}}(c, q)$ and $T_{\text{end}}(c, q)$, respectively.

\subsection{Bi-encoder Training}
\label{sec:biencoder}
Finally, we train a bi-encoder model to match the cross-encoder predictions on the generated questions.
This objective encourages the contextual representation for a token to have high similarity (in inner product space) with the representation of every question that is answered by that token.

\paragraph{Model.}
The bi-encoder model with parameters $\theta$ consists of of three components: an encoder $r$ and two question embedding heads $h_{\text{start}}$ and $h_{\text{end}}$ that map $\mathbb{R}^d \to \mathbb{R}^d$.
These heads will only be applied to beginning-of-sequence (i.e., \texttt{CLS}) representations;
as shorthand, define $f_{\text{start}}(x) = h_{\text{start}}(r(x)_1)$
and likewise for $f_{\text{end}}$.
Given a context passage $c$ and question $q$, the model predicts 
\begin{align}
p_{\theta}(a_{\text{start}} &= i \mid c, q) \propto e^{r(c)_i^\top f_{\text{start}}(q)} \\
p_{\theta}(a_{\text{end}} &= i \mid c, q) \propto e^{r(c)_i^\top f_{\text{end}}(q)}
\end{align}
In other words, the model independently encodes the passage and question with $r$,
applies the start and end heads to the \texttt{CLS} token embedding for $q$,
then predicts the answer start (end) index with a softmax over the dot product between the passage representation at that index and the output of the start (end) head.
We initialize $r$ to be the pre-trained RoBERTa-large model \citep{liu2019roberta}, which uses $d=1024$.
$h_{\text{start}}$ and $h_{\text{end}}$ are randomly-initialized 2-layer MLPs with hidden dimension $1024$, matching the default initialization of classification heads in RoBERTa.\footnote{\url{https://github.com/pytorch/fairseq/blob/master/fairseq/models/roberta/model.py}}

\paragraph{Training.} For an input consisting of context $c$ of length $L$ and question $q$, we train $\theta$ to minimize the KL-divergence between the student and teacher predictions, which is equivalent to the objective 
\begin{align}
- \sum_{i=1}^L & T_{\text{start}}(c, q)_i \log p_{\theta}(a_{\text{start}} = i \mid c, q)  \nonumber \\
& + T_{\text{end}}(c, q)_i \log p_{\theta}(a_{\text{end}} = i \mid c, q)
\end{align}
up to constants that do not depend on $\theta$.
We train for two epochs on the 80 million generated questions, which takes roughly 56 hours on 8 V100 GPUs, or roughly 19 GPU-days.\footnote{For comparison, pre-training RoBERTa-large from scratch took roughly 5000 GPU-days \citep{liu2019roberta}}
For efficiency, we process all questions for the same passage in the same batch, as encoding passages dominates runtime.
For further details, see Appendix~\ref{app:qadp-details}.

\section{Downstream Tasks}
\label{sec:downstream}
We evaluate \quip{} on zero-shot paraphrase ranking, few-shot paraphrase classification, few-shot NER, and zero-shot sentiment analysis.
Different tasks showcase different advantages of \quip{}.
For paraphrase detection and NER, \quip{} succeeds by learning meaningful token-level contextualized representations for single passages, whereas MLM representations are trained to reconstruct uncontextualized word embeddings, and the cross-encoder QA model is trained to represent passage-question pairs.
For NER and sentiment analysis, we prompt \quip{} with questions, leveraging its question-answering abilities.
Compared with a cross-encoder, \quip's bi-encoder architecture enables a more efficient way to use question prompts in NER, and yields more reliable scores when questions are unanswerable in sentiment analysis.
We focus on zero-shot and few-shot settings, as these require pre-trained models that are useful without fine-tuning on a large task-specific training dataset. 
\quip{} addresses this need by anticipating what information might be useful for downstream tasks---namely, information found in question-answer pairs.

\subsection{Paraphrase Ranking}
We first evaluate \quip{} token-level representations by measuring their usefulness for zero-shot paraphrase ranking.
In this task, systems must rank sentence pairs that are paraphrases above pairs that are non-paraphrases, without any task-specific training data.
We compute similarity scores using the $F_{\text{BERT}}$ variant of BERTScore \citep{zhang2020bertscore},
which measures cosine similarities between the representation of each token in one sentence and its most similar token in the other sentence. 
Given sentences $x_1$ and $x_2$ of lengths $L_1$ and $L_2$, define 
\begin{align*}
B(x_1, x_2) &= \frac{1}{L_1} \sum_{i=1}^{L_1} \max_{1 \le j \le L_2} \frac{r(x_1)_i^\top r(x_2)_j}{\|r(x_1)_i\|\|r(x_2)_j\|}.
\end{align*}
The $F_{\text{BERT}}$ BERTScore is defined as the harmonic mean of $B(x_1, x_2)$ and $B(x_2, x_1)$.
\citet{zhang2020bertscore} showed that BERTScore with RoBERTa is useful for both natural language generation evaluation and paraphrase ranking.
Since BERTScore uses token-level representations, we hypothesize that it should pair well with \quip.
As in \citet{zhang2020bertscore}, we use representations from the layer of the network that maximizes Pearson correlation between BERTScore and human judgments on the WMT16 metrics shared task \citep{bojar-etal-2016-results}.

\subsection{Paraphrase Classification}
We use either frozen or fine-tuned \quip{} representations for few-shot paraphrase classification, rather than ranking.
Through these experiments, we can compare \quip{} with existing work on few-shot paraphrase classification.

\paragraph{Frozen model.} 
We train a logistic regression model that uses BERTScore with frozen representations as features. For a given pair of sentences, we extract eight features, corresponding to BERTScore computed with the final eight layers (i.e., layers 17-24) of the network. These layers encompass the optimal layers for both RoBERTa-large and \quip{} (see \S\ref{sec:exp-paraphrase-ranking}).
Freezing the encoder is often useful in practice, particularly for large models, as the same model can be reused for many tasks \citep{brown2020language,du-etal-2020-general}.

\paragraph{Fine-tuning.}
For fine-tuning, we use the same computation graph and logistic loss function, but now backpropagate through the parameters of our encoder.
For details, see Appendix~\ref{app:finetune}.

\subsection{Named Entity Recognition}
\label{sec:ner}
We also use \quip{} for few-shot\footnote{
\citet{yang-katiyar-2020-simple} study few-shot NER assuming data from other NER datasets is available;
we assume no such data is available, matching \citet{huang2020few}.} named entity recognition, which we frame as a BIO tagging task.
Since questions in QA often ask for entities of a specific type, we expect \quip{} representations to contain rich entity type information.
We add a linear layer that takes in token-level representations and predicts the tag for each token, and backpropagate log loss through the entire network.
By default, the output layer is initialized randomly.

As a refinement, we propose using question prompts to initialize this model.
The output layer is parameterized by a $T \times d$ matrix $M$, where $T$ is the number of distinct BIO tags.
The log-probability of predicting the $j$-th tag for token $i$ is proportional to the dot product between the representation for token $i$ and the $j$-th row of $M$; this resembles how the bi-encoder predicts answers.
Thus, we initialize each row of $M$ with the start head embedding of a question related to that row's corresponding entity tag. For instance, we initialize the parameters for the \texttt{B-location} and \texttt{I-location} tags with the embedding for \nl{What is a location ?}
We normalize the question embeddings to have unit L2 norm.
This style of initialization is uniquely enabled by our bi-encoder QA model, as it builds a \emph{single} passage representation that can simultaneously answer questions corresponding to all entity types.
It would be unclear how to use a language model or a cross-encoder QA model similarly, as it must perform a separate forward pass for each question (i.e., each entity type in this setting).

\subsection{Zero-shot Sentiment Analysis}
\label{sec:sentiment}

Finally, we use \quip{} for zero-shot binary sentiment analysis.
We reduce sentiment analysis to QA by writing a pair of questions that ask for a \emph{reason} why an item is good or bad (e.g., \nl{Why is this movie [good/bad]?}). 
We predict the label whose corresponding question has higher similarity with the \quip{} representation of some token in the input.
This prompting strategy has the additional benefit of extracting rationales, namely the span that the \quip{} model predicts as the answer to the question.
While we focus on sentiment analysis, extractive rationales have been used for a wide range of NLP tasks \citep{deyoung-etal-2020-eraser}, suggesting that this method could be applied more broadly.

More formally, let $x$ be an input sentence and $(q_0, q_1)$ be a pair of questions (i.e., a prompt).
For label $y \in \{0, 1\}$, we compute a score for $y$ as 
\begin{align}
S(x, y) = &\max_{i} r(x)_i^\top f_{\text{start}}(q_y) + \nonumber \\
&\max_{i} r(x)_i^\top f_{\text{end}}(q_y).
\end{align}
This formula is a straightforward way to measure the extent to which \emph{some} span in $x$ looks like the answer to the question $q_y$, based on the model's pre-trained ability to perform QA.
We predict whichever $y$ has the higher value of $S(x, y) - C_y$, where $C_y$ is a calibration constant that offsets the model's bias towards answering $q_0$ or $q_1$.
Our inclusion of $C_y$ is inspired by
\citet{zhao2021calibrate}, who recommend calibrating zero-shot and few-shot models with a baseline derived from content-free inputs to account for biases towards a particular label.
To choose $C_y$,
we obtain a list $W$ of the ten most frequent English words, all of which convey no sentiment, 
and define $C_y$ as the mean over $w \in W$ of $S(w, y)$, i.e., the score when using $w$ as the input sentence
(see Appendix~\ref{app:calibration}).

This method can succeed only if the model produces a lower score for unanswerable questions than answerable ones.
For example, if the input passage is positive, the model must produce a lower score for \nl{Why is it bad?}, which not answerable (as the question contains a presupposition failure), than \nl{Why is it good?}, which presumably can be answered from the passage.
We hypothesize that \quip{} will indeed recognize that unanswerable questions should receive lower scores, as it is trained to make each span's representation far away from those of questions it does not answer.
In contrast, the cross-encoder objective does not teach the model how to handle unanswerable questions.

%% file: experiments.tex
\section{Experiments}
\subsection{Experimental details}
\paragraph{Datasets.} For paraphrasing, we use four datasets: QQP \citep{iyer2017qqp}, MRPC \citep{dolan-brockett-2005-automatically}, PAWS-Wiki, and PAWS-QQP \citep{zhang-etal-2019-paws}.
The PAWS datasets were designed to be challenging for bag-of-words models, and thus test whether our representations are truly contextual or mostly lexical.
For QQP and MRPC, we use the few-shot splits from \citet{gao2021making} that include 16 examples per class; for the PAWS datasets, we create new few-shot splits in the same manner.
We report results on the development sets of QQP and MRPC (as test labels were not available), the test set of PAWS-Wiki, and the ``dev-and-test'' set of PAWS-QQP.
For NER, we use two datasets: CoNLL 2003 \citep{tjong-kim-sang-de-meulder-2003-introduction} and WNUT-17 \citep{derczynski-etal-2017-results}.
We use the few-shot splits from \citet{huang2020few} that include 5 examples per entity type.
All few-shot experiments report an average over five random splits and seeds, following both \citet{gao2021making} and \citet{huang2020few}.
For sentiment analysis, we use two movie review datasets, SST-2 \citep{socher-etal-2013-recursive} and Movie Reviews (MR; \citealp{pang-lee-2005-seeing}), as well as the Customer Reviews (CR) dataset \citep{hu2004mining}.
We evaluate on the SST-2 development set and the MR and CR test sets made by \citet{gao2021making}.

\paragraph{Hyperparameter and prompt selection.} Due to the nature of zero-shot and few-shot experiments, we minimize the extent to which we tune hyperparameters,
relying on existing defaults and previously published hyperparameters.
For few-shot paraphrase classification, NER, and sentiment analysis, we developed our final method only using QQP, CoNLL, and SST-2, respectively, and directly applied it to the other datasets with no further tuning.
We did measure zero-shot paraphrase ranking accuracy on all datasets during development of \quip.
For more details, see Appendix~\ref{app:hyperparam}.

For NER, we used the first question prompts we wrote for both CoNLL and WNUT, which all follow the same format,
\nl{Who/What is a/an [entity type] ?} (see Appendix~\ref{app:ner-prompts} for all prompts).
For sentiment analysis, we wrote six prompts (shown in Appendix~\ref{app:sentiment-prompts}) and report mean accuracy over these prompts, to avoid pitfalls associated with prompt tuning \citep{perez2021true}. We use the same prompts for SST-2 and MR; for CR, the only change we make is replacing occurrences of the word \nl{movie} with \nl{product} to reflect the change in domain between these datasets.

\subsection{Baselines and Ablations}
\label{sec:ablations}
To confirm the importance of all three stages of our pre-training pipeline, we compare with a number of baselines and ablations.

\paragraph{No question generation.} We train the bi-encoder model directly on the MRQA training data (``\NoQgen''). We also include the cross-encoder teacher model trained on MRQA as a baseline (``\Teacher''). These settings mirror standard intermediate task training \citep{phang2018sentence,pruksachatkun-etal-2020-intermediate}.

\paragraph{No teacher.} We train the bi-encoder using the answer generated by the question generation model (``\NoTeacher''). If the generated answer is not a span in the passage, we consider the question unanswerable and treat the span containing the \texttt{CLS} token as the answer, as in \citet{devlin-etal-2019-bert}. 

\paragraph{Cross-encoder self-training.} To test whether the bottleneck imposed by the bi-encoder architecture is crucial for \quip, we also train a cross-encoder model on our generated data (``\CrossStudent''). Since this student model has the same architecture as the teacher model, we train it to match the teacher's argmax predictions, a standard self-training objective \citep{lee2013pseudo,kumar2020understanding}.
Training is much less efficient for the cross-encoder than the bi-encoder, since batching questions about the same passage together does not speed up training, so we train for a comparable number of GPU-hours (60 hours on 8 V100 GPUs).

\paragraph{Unsupervised QA.}  We test whether \quip{} requires real QA data, or if 
a rough approximation suffices.
We thus train a bi-encoder on 80 million pseudo-questions generated by applying noise to sentences (``\Unsup''),
as in \citet{lewis-etal-2019-unsupervised}.

\begin{table}[t]
\centering
\footnotesize
\begin{tabular}{l|cc}
\toprule
Model & EM & F1 \\
\midrule
\citet{lee2021learning} & $78.3$ & $86.3$ \\
\Unsup & $17.4$ & $24.9$ \\
\NoQgen & $70.7$ & $79.4$\\
\NoTeacher & $75.3$ & $84.7$ \\
\quip & $\bf 85.2$ & $\bf 91.7$ \\
\midrule
BERT-large cross-encoder & $84.2$ & $91.1$ \\
\Teacher & $88.8$ & $94.7$ \\
\CrossStudent & $\bf 89.5$ & $\bf 94.8$ \\
\bottomrule
\end{tabular}
\caption{EM and F1 scores on the SQuAD development set for bi-encoder (top) and cross-encoder (bottom) models.
\quip{} outperforms the other bi-encoder model baselines, and even a cross-encoder BERT-large model.
The RoBERTa cross-encoder models are better at QA, but will underperform \quip{} on non-QA tasks.
}
\label{tab:qa} 
\end{table}

\subsection{Bi-encoder Question Answering}
\label{sec:exp-qa}

While not our main focus, we first check that \quip{} improves bi-encoder QA accuracy, as shown in Table~\ref{tab:qa}.
\quip{} improves over \citet{lee2021learning} by $5.4$ F1 on the SQuAD development set. 
It also surpasses the reported human accuracy of $91.2$ F1 on the SQuAD test set, as well as the best cross-encoder BERT-large single model from \citet{devlin-etal-2019-bert}.
\quip{} greatly improves over baselines that directly train on MRQA data or do not use the teacher model.
The cross-encoder models are more accurate at QA, but as we will show, this does not imply that cross-encoder QA is a better pre-training objective for downstream non-QA tasks.
Appendix~\ref{app:qa} shows results on all MRQA development datasets.

\subsection{Zero-shot Paraphrase Ranking}
\label{sec:exp-paraphrase-ranking}

\begin{table*}[t]
\centering
\footnotesize
\begin{tabular}{l|cc|cccc}
\toprule
Model & WMT $r$ & WMT Best Layer & QQP & MRPC & PAWS-Wiki & PAWS-QQP \\
\midrule
RoBERTa-large & $.739$ & $17$ & $.763$ & $.831$ & $.698$ & $.690$ \\
\Teacher & $.744$ & $16$ & $.767 $ & $.840$ & $.742$ & $.731$ \\
\CrossStudent & $.753$ & $16$ & $.769$ & $.847$ & $.751$ & $.706$ \\
\Unsup & $.654$ & $11$ & $.747$ & $.801$ & $.649$ & $.580$ \\
\NoQgen & $.749$ & $15$ & $.771$ & $.807$ & $.747$ & $.725$ \\
\NoTeacher & $.726$ & $19$ & $.767$ & $.831$ & $.780$ & $.709$ \\
\quip & $\bf .764$ & $20$ & $\bf .809$ & $\bf .849$ & $ \bf .830$ & $\bf .796$ \\
\bottomrule
\end{tabular}
\caption{Pearson correlation on WMT development data, best layer chosen based on WMT results, and AUROC on zero-shot paraphrase ranking using BERTScore. \quip{} outperforms all baselines on all datasets.}
\label{tab:zeroshot-paraphrase} 
\end{table*}

We validate our approach and study the effects of various ablations on zero-shot paraphrase ranking.
The first half of Table~\ref{tab:zeroshot-paraphrase} shows WMT development set Pearson correlations averaged across six to-English datasets, as in \citet{zhang2020bertscore}, along with the best layer for each model.
\quip{} reaches its optimal score at a later layer (20) than RoBERTa-large (17), which may suggest that the \quip{} training objective is more closely aligned with learning better representations than MLM.  

The rest of Table~\ref{tab:zeroshot-paraphrase} shows zero-shot paraphrase ranking results using BERTScore.
\quip{} improves substantially over RoBERTa on all four datasets,
with an average improvement of $.076$ AUROC.
The improvement is greatest on the PAWS datasets; 
since these datasets cannot be solved by lexical features alone, \quip{} representations must be much more contextualized than RoBERTa representations. 
Training on Unsupervised QA data degrades performance compared to RoBERTa, showing that \quip{} does not merely make word representations encode local context in a simple way.
Training the bi-encoder directly on the MRQA dataset or without the teacher improves on average over RoBERTa, but \quip{} greatly outperforms both baselines.
The cross-encoder models also lag behind \quip{} at paraphrase ranking, despite their higher QA accuracy; since the cross-encoders are trained to take passage-question pairs as inputs, their representations of single sentences are not as useful.
Thus, we conclude that having real questions, accurate answer supervision, and a bi-encoder student model are all crucial to the success of \quip.

\subsection{Paraphrase Classification}
\label{sec:exp-paraphrase-classification}
\begin{table*}[t]
\centering
\footnotesize
\begin{tabular}{ll|cccc}
\toprule
Model & Fine-tuned? & QQP & MRPC & PAWS-Wiki & PAWS-QQP \\
\midrule
LM-BFF (reported) & Fine-tuned & $69.8_{0.8}$ & $77.8_{0.9}$ & - & - \\
LM-BFF (rerun) & Fine-tuned & $67.1_{0.9}$ & $76.5_{1.5}$ & $60.7_{0.7}$ & $50.1_{2.8}$ \\
\midrule
RoBERTa-large & Frozen & $64.4_{0.4}$ & $80.6_{0.7}$ & $62.3_{0.9}$ & $50.6_{0.4}$ \\
\quip & Frozen & $68.9_{0.2}$ &  $82.6_{0.4}$ & $71.9_{0.5}$ & ${\bf 63.0_{1.2}}$ \\
\midrule
RoBERTa-large & Fine-tuned & $64.9_{0.7}$ & $84.4_{0.3}$ & $65.7_{0.3}$ & $50.9_{0.8}$\\
\quip & Fine-tuned & ${\bf 71.0_{0.3}}$ & ${\bf 86.6_{0.4}}$ & ${\bf 75.1_{0.2}}$ & $60.9_{1.0}$ \\
\bottomrule
\end{tabular}
\caption{F1 scores on few-shot paraphrase classification, averaged across five training splits (standard errors in subscripts).
\quip{} outperforms prior work (LM-BFF; \citealp{gao2021making})
as well as our own RoBERTa baselines.
}
\label{tab:fewshot-paraphrase} 
\end{table*}

Table~\ref{tab:fewshot-paraphrase} shows few-shot paraphrase classification results.
As we studied \quip-related ablations in the previous section, we focus on the comparison between \quip{} and baselines based on MLM.
First, we use RoBERTa-large embeddings in place of \quip{} in our method. 
Second, we compare with LM-BFF \citep{gao2021making}, which pairs RoBERTa-large with MLM-style prompts.
We use LM-BFF with manually written prompts and demonstrations, which was their best method on QQP by $2.1$ F1 and was $0.3$ F1 worse than their best method on MRPC.
\quip{} used as a frozen encoder is competitive with LM-BFF on QQP and outperforms it by $6.1$ F1 on MRPC, $11.2$ F1 on PAWS-Wiki, and $12.1$ F1 on PAWS-QQP.
Fine-tuning \quip{} gives additional improvements on three of the four datasets,
and outperforms fine-tuning RoBERTa by an average of $6.9$ F1.

\subsection{Named Entity Recognition}
\label{sec:exp-ner}

\begin{table}[t]
\centering
\footnotesize
\begin{tabular}{l|cc}
\toprule
Model & CoNLL & WNUT \\
\midrule
\citet{huang2020few} & $65.4$ & $37.6$ \\
\midrule
\textbf{Standard init.} & & \\
RoBERTa-large & $59.0_{2.4}$ & $39.3_{0.6}$ \\
\Teacher & $68.9_{3.3}$ & $43.0_{0.9}$ \\
\CrossStudent & $63.4_{3.3}$ & $39.4_{1.7}$ \\
\Unsup & $58.2_{2.6}$ & $26.0_{1.0}$ \\
\NoQgen & $66.4_{3.3}$ & $42.2_{0.4}$ \\
\NoTeacher & $67.7_{1.9}$ & $40.7_{1.4}$ \\
\quip & $70.0_{2.4}$ & $42.2_{0.5}$ \\
\midrule
\textbf{Question prompt init.} & & \\
\Unsup & $62.7_{3.3}$ & $30.4_{0.8}$ \\
\NoQgen & $72.0_{2.8}$ & $44.0_{1.3}$ \\
\NoTeacher & $71.4_{3.0}$ & $47.8_{1.1}$ \\
\quip & $\bf 74.0_{2.4}$ & $\bf 49.6_{0.5}$ \\
\bottomrule
\end{tabular}
\caption{F1 scores on few-shot NER, averaged over five training splits (standard errors in subscripts).
\quip{} with question prompts performs best on both datasets.
}
\label{tab:few-ner} 
\end{table}

Table~\ref{tab:few-ner} shows few-shot NER results on the CoNLL and WNUT datasets. \quip{} improves over RoBERTa-large by $11$ F1 on CoNLL and $2.9$ F1 on WNUT when used with a randomly initialized output layer.
We see a further improvement of $4$ F1 on CoNLL and $7.4$ F1 on WNUT when using question embeddings to initialize the output layer. 
Using the cross-encoder trained directly on QA data is roughly as good as \quip{} when using randomly initialized output layers, but it is incompatible with question embedding initialization. 

\subsection{Sentiment Analysis}
\label{sec:exp-sentiment}

\begin{table}[t]
\centering
\footnotesize
\begin{tabular}{l|ccc}
\toprule
Model & SST-2 & MR & CR \\
\midrule
CC + GPT-3 & $71.6$ & - & - \\
LM-BFF & $83.6$ & $80.8$ & $79.5$ \\
\quip{} (average) & $\bf 87.9_{0.6}$ & $\bf 81.9_{0.4}$ & $\bf 90.3_{0.2}$ \\
w/ cross-enc. student & $83.3_{0.4}$ & $78.5_{0.4}$ & $88.9_{0.3}$ \\
\midrule
\quip{} (tune on SST-2) & $89.6$ & $83.1$ & $90.4$ \\
\bottomrule
\end{tabular}
\caption{Zero-shot accuracy on sentiment analysis. 
Third and fourth rows show mean accuracy across six prompts (standard error in subscripts).
\quip{} with an average prompt outperforms prior work;
using the best prompt on SST-2 helps on all datasets.
}
\label{tab:sentiment} 
\end{table}

\begin{table}[t]
\centering
\scriptsize
\begin{tabular}{c|p{2.4in}}
\toprule
Label & Rationale \\
\midrule
- & \nl{too slim}, \nl{stale}, \nl{every idea}, \nl{wore out its welcome}, \nl{unpleasant viewing experience}, \nl{lifeless}, \nl{plot}, \nl{amateurishly assembled}, \nl{10 times their natural size}, \nl{wrong turn} \\
\midrule
+ & \nl{packed with information and impressions}, \nl{slash-and-hack}, \nl{tightly organized efficiency}, \nl{passion and talent}, \nl{best films}, \nl{surprises}, \nl{great summer fun}, \nl{play equally well}, \nl{convictions}, \nl{wickedly subversive bent} \\
\bottomrule
\end{tabular}
\caption{Rationales extracted by \quip{} on ten random examples for each label from SST-2.}
\label{tab:example-rationales} 
\end{table}

Table~\ref{tab:sentiment} shows zero-shot accuracy on our three sentiment analysis datasets.
We compare with zero-shot results for LM-BFF \citep{gao2021making}\footnote{We tried applying our calibration strategy to LM-BFF as well, but found that it did not improve accuracy.}
and reported zero-shot results from \citet{zhao2021calibrate} using GPT-3 with Contextual Calibration (CC) on SST-2.
\quip{} using an average prompt outperforms zero-shot LM-BFF by $5.4$ points, averaged across the three datasets.
Choosing the best prompt on SST-2 and using that for all datasets improves results not only on SST-2 but also MR, and maintains average accuracy on CR.
Using the cross-encoder student QA model with the same prompts leads to worse performance: 
we hypothesize that the bi-encoder succeeds due to its better handling of unanswerable questions.
Overall, these results show that question answering can provide a viable interface for building models that perform non-QA tasks.

Table~\ref{tab:example-rationales} shows rationales extracted from random SST-2 examples for which \quip{} was correct with the best prompt for SST-2 (\nl{What is the reason this movie is [good/bad]?}).
To prefer shorter rationales, we extract the highest-scoring span of five BPE tokens or less.
The model often identifies phrases that convey clear sentiment.
Appendix~\ref{app:sentiment-rationales} shows full examples and rationales. 

\subsection{Stability Analysis}
\label{sec:stability}
We experimented with some design decisions that did not materially affect our results. 
Appendix~\ref{app:stability} shows results for three such choices: 
including in-batch negative passages \citep{lee2021learning}, using the argmax prediction of the teacher rather than soft labels, and using beam search to generate a diverse set of answers followed by one high-likelihood question per answer.
We take these findings as evidence that our basic recipe is stable to many small changes.
For question generation, we hypothesize that the objective of matching the cross-encoder teacher model encourages the bi-encoder to learn important features identified by the cross-encoder, even on questions that are not entirely well-formed.

%% file: discussion.tex
\section{Discussion and Related Work}
We build on work in question generation and answering, pre-training, and few-shot learning.

\subsection{Question Generation}
Neural question generation has been well-studied for different purposes \citep{du-etal-2017-learning,du-cardie-2018-harvesting,zhao-etal-2018-paragraph,lewis2019generative,alberti-etal-2019-synthetic,puri-etal-2020-training,lewis2021paq,bartolo2021improving}.
We use generated questions to learn general-purpose representations.
We also show that a relatively simple strategy of generating the answer and question together with a single model can be effective; most prior work uses separate answer selection and question generation models.

\paragraph{Phrase-indexed Question Answering}
Phrase-indexed question answering is a paradigm for open-domain QA that retrieves answers by embedding questions and candidate answers in a shared embedding space \citep{seo-etal-2018-phrase,seo-etal-2019-real,lee-etal-2020-contextualized}.
It requires using a bi-encoder architecture for efficient phrase retrieval.
Especially related is \citet{lee2021learning}, 
which also uses question generation and a cross-encoder teacher model to improve phrase-indexed QA,
though they focus on improving QA accuracy rather than transfer to other tasks.
Our results reinforce prior observations that bi-encoder models are usually less accurate at QA than cross-encoders (see Table~\ref{tab:qa}).
However, the bi-encoder model transfers better to settings that require a contextualized representation of a single passage; the cross-encoder instead optimizes for producing representations of passage-question pairs.

\subsection{Improving question answering}
While we use QA to aid pre-training, related work aims to improve accuracy on QA.
\citet{ram2021fewshot} propose a span extraction pre-training objective that enables few-shot QA. 
\citet{khashabi-etal-2020-unifiedqa} run multi-task training on many QA datasets, both extractive and non-extractive, to improve QA accuracy.

\subsection{Learning contextual representations}
Pre-training on unlabeled data has yields useful contextual representations \citep{peters-etal-2018-deep,devlin-etal-2019-bert},
but further improvements are possible using labeled data.
Intermediate task training \citep{phang2018sentence} improves representations by training directly on large labeled datasets.
Muppet \citep{aghajanyan2021muppet} improves models by multi-task pre-finetuning on many labeled datasets.

Most similar to our work, QuASE \citep{he-etal-2020-quase} uses extractive QA to pre-train a BERT paragraph encoder.
Our work improves upon QuASE in multiple ways. 
First, we use question generation and knowledge distillation to greatly improve over directly training on labeled data, the approach used by QuASE.
Second, we propose multiple ways of leveraging question-based task descriptions to improve accuracy in zero-shot and few-shot settings, thus showing how the QA format can be used as a model-building interface for non-QA tasks;
QuASE only uses their model as a feature extractor.
Moreover, since the architecture of QuASE involves a more complex interaction layer than our bi-encoder, it would not be possible to use question prompts to initialize final-layer parameters, as we do for NER.

Other work has used methods similar to ours to learn vector representations of full sentences.
\citet{reimers-gurevych-2019-sentence} train sentence embeddings for sentence similarity tasks using natural language inference data.
\citet{thakur-etal-2021-augmented} train a sentence embedding bi-encoder to mimic the predictions of a cross-encoder model.
We learn token-level representations,
rather than a single vector for a sentence, and thus use token-level supervision from extractive QA.

\subsection{Few-shot learning}
We study few-shot learning without access to unlabeled data, following most recent work \citep{brown2020language,gao2021making,zhao2021calibrate}.
\citet{schick-schutze-2021-just} notably propose a semi-supervised approach that uses unlabeled data for knowledge distillation; this process does not improve accuracy, but mainly improves efficiency. 
Moreover, large-scale unlabeled data may not be easily obtainable for all tasks, and utilizing such data increase computation time in the fine-tuning stage, so we focus on the setting without unlabeled data.
The aforementioned work uses language models for few-shot learning by converting tasks to language modeling problems; we develop alternative methods for few-shot learning that use token-level representations and question-based prompts.

\section{Conclusion}
In this work, we pre-trained token-level contextual representations that are useful for downstream few-shot learning.
Our key idea was to use question-answer pairs to define what information should be encoded in passage representations.
We showed that these representations are useful for a variety of standard NLP tasks in zero- and few-shot settings, including paraphrase detection, named entity recognition, and sentiment analysis, across nine total datasets.
Looking forward, we hope to see more work on designing pre-training objectives that align with downstream needs for few-shot learning.

%% file: appendix.tex
\section{Appendix}
\label{sec:appendix}

\subsection{\quip{} Details}
\label{app:qadp-details}
We limit passages to $456$ byte-pair encoding (BPE) tokens and questions to $50$ so that the concatenation can fit comfortably within the $512$ token context usable by the cross-encoder teacher.
We create passages from our unlabeled text corpus by greedily selecting maximal chunks of contiguous sentences that fit within the BPE token limit.
We pre-compute the teacher predictions $T_{\text{start}}$ and $T_{\text{end}}$ before bi-encoder training. To save space, we sparsify these vectors by only storing the eight largest predicted probabilities, treating all others as $0$.

We conducted minimal hyperparameter tuning for \quip.
We used a learning rate of $1 \cdot 10^{-5}$ (default for most RoBERTa fine-tuning experiments\footnote{\url{https://github.com/pytorch/fairseq/tree/master/examples/roberta}}) and no gradient accumulation, which we found led to faster training.

\subsection{Paraphrase Fine-tuning Details}
\label{app:finetune}
To fine-tune our model for paraphrase classification,
we use two practices recommended by  \citet{mussmann-etal-2020-importance}, who also train a binary classification model that uses cosine similarity-based features derived from fine-tuned BERT embeddings.
First, we disable dropout during training, as dropout artificially lowers all cosine similarities. 
Second, we use a larger learning rate on the final output layer than the Transformer parameters, by a factor of $10^3$.

\subsection{Downstream Task Hyperparameter Details}
\label{app:hyperparam}
For few-shot paraphrase detection with the frozen model, we use Scikit-learn's logistic regression implementation with default settings \citep{scikit-learn}.
For fine-tuned paraphrase detection, we again use a learning rate of $1 \cdot 10^{-5}$ and train for $20$ epochs, which we found to usually be sufficient for convergence on the training data.
For NER, we use the default hyperparameters from the Huggingface \texttt{transformers} repository \citep{wolf-etal-2020-transformers},
with the exception of decreasing the learning rate from $5 \cdot 10^{-5}$ to $2 \cdot 10^{-5}$, which we found improved the RoBERTa baseline on CoNLL.

\subsection{Sentiment Analysis Calibration}
\label{app:calibration}
To calibrate the zero-shot sentiment analysis model, we use ten content-free inputs: \nl{the}, \nl{be}, \nl{to}, \nl{of}, \nl{and}, \nl{a}, \nl{in}, \nl{that}, \nl{have}, and \nl{I}.
These were the top ten words listed on 
\url{https://en.wikipedia.org/wiki/Most_common_words_in_English}.
We only applied calibration for the main \quip{} model, as we did not find calibration to improve results for either LM-BFF or the cross-encoder QA student model.

\subsection{Full QA results}
\label{app:qa}
Table~\ref{tab:mrqa} shows EM and F1 scores on the 12 development sets from the MRQA 2019 Shared Task \citep{fisch-etal-2019-mrqa}.
These are divided into 6 in-domain datasets---HotpotQA \citep{yang-etal-2018-hotpotqa}, NaturalQuestions \citep{kwiatkowski-etal-2019-natural}, NewsQA \citep{trischler-etal-2017-newsqa}, SearchQA \citep{dunn2017searchqa}, SQuAD \citep{rajpurkar-etal-2016-squad}, and TriviaQA \citep{joshi-etal-2017-triviaqa}---for which corresponding training data was used to train the question generation model and teacher,
and 6 out-of-domain datasets---BioASQ \citep{tsatsaronis2015bioasq}, DROP \citep{dua-etal-2019-drop}, DuoRC \citep{saha-etal-2018-duorc}, RACE \citep{lai-etal-2017-race}, RelationExtraction \citep{levy-etal-2017-zero}, and TextbookQA \citep{kembhavi2017textbook}---for which no training data was used in the \quip{} pipeline.
\quip{} improves over training the bi-encoder directly on the MRQA data by an average of $4.4$ F1 on the in-domain datasets and $12.7$ F1 on the out-of-domain datasets.
It underperforms the cross-encoder teacher by about $5$ F1 on both the in-domain and out-of-domain datasets on average.
\begin{table*}[t]
\centering
\scriptsize
\begin{tabular}{l|ccccccc}
\toprule
In-domain & HotpotQA & NaturalQ & NewsQA & SQuAD & SearchQA & TriviaQA & Average \\
\midrule
\Unsup &  9.5 / 16.6 & 8.0 / 15.5 & 7.6 / 14.4 & 17.5 / 25.0 & 15.4 / 21.1 & 17.6 / 23.3 & 12.6 / 19.3 \\
\NoQgen & 61.0 / 77.5 & 64.1 / 76.4 & 46.1 / 61.5 & 70.9 / 79.6 & 73.8 / 79.8 & 63.1 / 69.0 & 63.2 / 74.0 \\
\NoTeacher & 52.9 / 68.7 & 57.8 / 70.8 & 41.8 / 58.7 & 75.4 / 84.8 & 64.5 / 71.7 & 71.1 / 76.1 & 60.6 / 71.8 \\
\quip & 61.3 / 77.9 & 63.7 / 77.2 & 52.4 / 68.7 & 85.3 / 91.8 & 68.7 / 76.8 & 72.0 / 78.1 & 67.2 / 78.4 \\
\midrule
\Teacher & 66.8 / 83.0 & 70.5 / 82.0 & 58.8 / 72.9 & 89.1 / 94.8 & 78.3 / 84.6 & 73.4 / 79.6 & 72.8 / 82.8 \\
\CrossStudent & 66.3 / 82.3 & 66.5 / 79.4 & 54.4 / 70.5 & 89.6 / 94.9 & 72.1 / 80.1 & 73.4 / 79.8 & 70.4 / 81.2 \\
\bottomrule
\toprule
Out-of-domain & BioASQ & DROP & DuoRC & RACE & RelationExt & TextbookQA & Average \\
\midrule
\Unsup & 15.3 / 19.2 & 5.9 / 9.5 & 14.1 / 17.4 & 6.5 / 11.4 & 12.7 / 22.1 & 8.9 / 13.3 & 10.6 / 15.5 \\
\NoQgen & 42.2 / 57.2 & 29.9 / 38.3 & 38.6 / 48.6 & 29.1 / 39.8 & 71.3 / 83.5 & 34.7 / 43.6 & 41.0 / 51.8 \\
\NoTeacher & 40.9 / 54.9 & 33.5 / 43.0 & 44.1 / 53.2 & 31.8 / 44.4 & 70.8 / 82.1 & 37.3 / 46.2 & 43.0 / 54.0 \\
\quip & 51.3 / 67.5 & 46.2 / 57.1 & 53.0 / 63.2 & 39.6 / 53.4 & 75.5 / 86.0 & 50.2 / 60.0 & 52.6 / 64.5 \\
\midrule
\Teacher & 58.0 / 72.9 & 55.4 / 65.3 & 55.0 / 66.8 & 44.2 / 57.7 & 78.5 / 88.8 & 58.5 / 67.4 & 58.2 / 69.8 \\
\CrossStudent & 57.3 / 72.6 & 57.5 / 68.3 & 56.2 / 67.5 & 44.8 / 58.6 & 79.5 / 89.1 & 58.4 / 67.3 & 59.0 / 70.6 \\
\bottomrule
\end{tabular}
\caption{Exact match/F1 scores on the twelve development datasets from the MRQA 2019 shared task.
The six in-domain datasets are on top; the six out-of-domain datasets are on bottom.}
\label{tab:mrqa} 
\end{table*}

\subsection{Stability Analysis}
\label{app:stability}

\begin{table}[t]
\centering
\footnotesize
\begin{tabular}{l|ccc}
\toprule
\multirow{2}{*}{Model} & SQuAD & Paraphrase & NER \\
& F1 & AUROC & F1 \\
\midrule
\quip & $91.7$ & $.821$ & $61.8$ \\
+ concat. passages & $91.7$ & $.818$ & $62.7$ \\
w/ hard labels & $91.5$ & $.814$ & $62.5$ \\
w/ 2-stage beam search & $91.7$ & $.821$ & $62.8$ \\
\bottomrule
\end{tabular}
\caption{SQuAD development set F1, average zero-shot paraphrase ranking AUROC across all datasets, and average few-shot NER F1 using question prompts across both datasets for \quip{} variants. Models shown here are all similarly effective.} 
\label{tab:stability} 
\end{table}

We experimented with some design decisions that did not materially affect our results.
Here, we report these findings as evidence that our basic recipe is stable to many small changes.
First, we concatenated the representations of all passages in the same batch and on the same GPU together (9 passages on average),
and trained the model to extract answers from this larger pseudo-document; this effectively adds in-batch negative passages, as in \citet{lee2021learning}.
Second, we trained the model to match the argmax prediction of the teacher, rather than its soft distribution over start and end indices.
Finally, we used a two-stage beam search to generate questions. For a given passage, we generated 20 possible answers via beam search, chose 10 of these to maximize answer diversity, then generated one question for each answer with another beam search. 
Our goal was to ensure diversity by forcing questions to be about different answers, while also maintaining high question quality.
As shown in Table~\ref{tab:stability}, these choices have a relatively minor impact on the results (within $.007$ AUROC and $1$ F1 on NER).

\subsection{QA Prompts for NER}
\label{app:ner-prompts}
Table~\ref{tab:ner-prompts} shows the question prompts we use to initialize the NER model for CoNLL and WNUT. For entity types that occur in both datasets, and for the \texttt{O} tag, we always use the same question.
We used the English description of the entity type provided by the dataset.

\begin{table}[t]
\centering
\small
\begin{tabular}{l|l}
\toprule
Entity type & Question \\
\midrule
\textbf{Both datasets} & \\
O & \nl{What is a generic object ?} \\
Person & \nl{Who is a person ?} \\
Location & \nl{What is a location ?} \\
\midrule
\textbf{CoNLL} & \\
Organization & \nl{What is an organization ?} \\
Miscellaneous & \nl{What is a miscellaneous entity ?} \\
\midrule
\textbf{WNUT} & \\
Corporation & \nl{What is a corporation ?} \\
Product & \nl{What is a product ?} \\
Creative work &  \nl{What is a creative work ?} \\
Group & \nl{What is a group ?} \\
\bottomrule
\end{tabular}
\caption{Question prompts used for the CoNLL and WNUT NER datasets. }
\label{tab:ner-prompts} 
\end{table}

\subsection{Full training set NER}
Table~\ref{tab:many-ner} shows NER results when training on the full training dataset. \quip{} gives a $0.6$ F1 improvement on WNUT, but has effectively the same accuracy on CoNLL. 
\label{app:full-ner}
\begin{table}[t]
\centering
\small
\begin{tabular}{l|cc}
\toprule
Model & CoNLL & WNUT \\
\midrule
RoBERTa-large & $92.7$ & $57.9$ \\
\quip, standard & $92.7$ & $58.1$ \\
\quip, QA prompts & $92.8$ & $58.8$ \\
\bottomrule
\end{tabular}
\caption{F1 scores on NER, using the entire training dataset.}
\label{tab:many-ner} 
\end{table}

\subsection{Sentiment Analysis QA Prompts}
\label{app:sentiment-prompts}
Table~\ref{tab:sentiment-prompts} shows the six prompts we use for sentiment analysis for the movie review datasets (SST-2 and MR). Each prompt consists of one question for the positive label and one for the negative label.
For CR, we use the same prompts except that we replace all instances of the word \nl{movie} with \nl{product}.

\begin{table}[t]
\centering
\small
\begin{tabular}{c|c|l}
\toprule
\# & Label & Question \\
\midrule
\multirow{2}{*}{1} & + & \nl{Why is it good?} \\
& - & \nl{Why is it bad?} \\
\midrule
\multirow{2}{*}{2}  & + & \nl{Why is this movie good?} \\
& - & \nl{Why is this movie bad?} \\
\midrule
\multirow{2}{*}{3}  & + & \nl{Why is it great?} \\
& - & \nl{Why is it terrible?} \\
\midrule
\multirow{2}{*}{4}  & + & \nl{What makes this movie good?} \\
& - & \nl{What makes this movie bad?} \\
\midrule
\multirow{2}{*}{5}  & + & \nl{What is the reason this movie is good?} \\
& - & \nl{What is the reason this movie is bad?} \\
\midrule
\multirow{2}{*}{6}  & + & \nl{What is the reason this movie is great?} \\
& - & \nl{What is the reason this movie is terrible?} \\
\bottomrule
\end{tabular}
\caption{Question prompts used for sentiment analysis on movie review datasets (SST-2 and MR). Prompts used for CR are identical except for replacing \nl{movie} with \nl{product}. }
\label{tab:sentiment-prompts} 
\end{table}

\subsection{Sentiment Analysis Rationales}
Tables~\ref{tab:sst2-rationales}, \ref{tab:mr-rationales}, and \ref{tab:cr-rationales} show full examples and rationales extracted by our zero-shot sentiment analysis method for SST-2, MR, and CR, respectively.
In all cases, we use the prompt that led to the highest accuracy on SST-2.
For each dataset, we randomly sample ten examples of each label for which the model predicted the correct answer.
We highlight in bold the span of $\le 5$ BPE tokens that the model predicts best answers the question associated with the correct label.
In some cases, 
the rationales correspond to clear sentiment markers.
In other cases, they highlight an aspect of a movie or product that is criticized or praised in the review; these could be considered reasonable answers to a question like \nl{Why is this movie bad?} even if the sentiment associated with them is unclear without the surrounding context.
In future work, it would be interesting to find better ways to align the task of extractive QA and with the goal of producing rationales that are human-interpretable in isolation.

\label{app:sentiment-rationales}
\begin{table*}[t]
\centering
\small
\begin{tabular}{c|p{5.5in}}
\toprule
Label & SST-2 Example (rationale in bold) \\
\midrule
\multirow{10}{*}{-} & \nl{for starters , the story is just \textbf{too slim} .} \\
& \nl{paid in full is so \textbf{stale} , in fact , that its most vibrant scene is one that uses clips from brian de palma 's scarface .} \\
& \nl{( e ) ventually , \textbf{every idea} in this film is flushed down the latrine of heroism .} \\
& \nl{corpus collosum -- while undeniably interesting -- \textbf{wore out its welcome} well before the end credits rolled about 45 minutes in .} \\
& \nl{makes for a pretty \textbf{unpleasant viewing experience} .} \\
& \nl{while ( hill ) has learned new tricks , the tricks alone are not enough to salvage this \textbf{lifeless} boxing film .} \\
& \nl{it 's hampered by a lifetime-channel kind of \textbf{plot} and a lead actress who is out of her depth .} \\
& \nl{dull , lifeless , and \textbf{amateurishly assembled} .} \\
& \nl{the movie is what happens when you blow up small potatoes to \textbf{10 times their natural size} , and it ai n't pretty .} \\
& \nl{every time you look , sweet home alabama is taking another bummer of a \textbf{wrong turn} .} \\
\midrule
\multirow{10}{*}{+} & \nl{though only 60 minutes long , the film is \textbf{packed with information and impressions} .} \\
& \nl{good old-fashioned \textbf{slash-and-hack} is back !} \\
& \nl{with \textbf{tightly organized efficiency} , numerous flashbacks and a constant edge of tension , miller 's film is one of 2002 's involvingly adult surprises .} \\
& \nl{displaying about equal amounts of naiveté , \textbf{passion and talent} , beneath clouds establishes sen as a filmmaker of considerable potential .} \\
& \nl{` easily my choice for one of the year 's \textbf{best films} . '} \\
& \nl{a delectable and intriguing thriller filled with \textbf{surprises} , read my lips is an original .} \\
& \nl{it is \textbf{great summer fun} to watch arnold and his buddy gerald bounce off a quirky cast of characters .} \\
& \nl{the film will \textbf{play equally well} on both the standard and giant screens .} \\
& \nl{for this reason and this reason only -- the power of its own steadfast , hoity-toity \textbf{convictions} -- chelsea walls deserves a medal .} \\
& \nl{there 's a \textbf{wickedly subversive bent} to the best parts of birthday girl .} \\
\bottomrule
\end{tabular}
\caption{Rationales (in bold) extracted by the zero-shot \quip{} sentiment analysis model for SST-2.
We show ten random examples for each label on which the model made the correct prediction.}
\label{tab:sst2-rationales} 
\end{table*}

\begin{table*}[t]
\centering
\small
\begin{tabular}{c|p{5.5in}}
\toprule
Label & MR Example (rationale in bold) \\
\midrule
\multirow{10}{*}{-} & \nl{strangely comes off as a kingdom \textbf{more mild than wild} .} \\
& \nl{feels like the work of someone who may indeed have finally \textbf{aged past his prime} . . . and , perhaps more than he realizes , just wants to be liked by the people who can still give him work .} \\
& \nl{watching the powerpuff girls movie , my mind kept returning to one anecdote for comparison : the cartoon in japan that \textbf{gave people seizures} .} \\
& \nl{this is a movie so insecure about its \textbf{capacity to excite} that it churns up not one but two flagrantly fake thunderstorms to underscore the action .} \\
& \nl{\textbf{witless} , pointless , tasteless and idiotic .} \\
& \nl{the next big thing's not-so-big ( and \textbf{not-so-hot} ) directorial debut .} \\
& \nl{unfortunately , it's also \textbf{not very good} . especially compared with the television series that inspired the movie .} \\
& \nl{\textbf{irwin and his director} never come up with an adequate reason why we should pay money for what we can get on television for free .} \\
& \nl{with this new rollerball , sense and sensibility have been overrun by what can only be characterized as \textbf{robotic sentiment} .} \\
& \nl{the video work is so \textbf{grainy and rough} , so dependent on being 'naturalistic' rather than carefully lit and set up , that it's exhausting to watch .} \\
\midrule
\multirow{10}{*}{+} & \nl{the \textbf{appearance of treebeard} and gollum's expanded role will either have you loving what you're seeing , or rolling your eyes . i loved it ! gollum's 'performance' is incredible !} \\
& \nl{droll \textbf{caper-comedy} remake of " big deal on madonna street " that's a sly , amusing , laugh-filled little gem in which the ultimate " bellini " begins to look like a " real kaputschnik . "} \\
& \nl{katz uses archival footage , horrifying documents of lynchings , still photographs and charming old reel-to-reel recordings of meeropol entertaining his children to create his song history , but most powerful of all is \textbf{the song itself}} \\
& \nl{a thunderous ride at first , quiet cadences of pure finesse are few and far between ; their shortage dilutes the potency of otherwise respectable action . still , this flick is \textbf{fun} , and host to some truly excellent sequences .} \\
& \nl{\textbf{compellingly watchable} .} \\
& \nl{an unbelievably fun film just a \textbf{leading man away from perfection} .} \\
& \nl{andersson creates a world that's at once \textbf{surreal and disturbingly familiar} ; absurd , yet tremendously sad .} \\
& \nl{the \textbf{invincible} werner herzog is alive and well and living in la} \\
& \nl{you can feel the heat that ignites this gripping tale , and the \textbf{humor and humanity} that root it in feeling .} \\
& \nl{this is a terrific \textbf{character study} , a probe into the life of a complex man .} \\
\bottomrule
\end{tabular}
\caption{Rationales (in bold) extracted by the zero-shot \quip{} sentiment analysis model for the Movie Reviews (MR) dataset.
We show ten random examples for each label on which the model made the correct prediction.}
\label{tab:mr-rationales} 
\end{table*}

\begin{table*}[t]
\centering
\small
\begin{tabular}{c|p{5.5in}}
\toprule
Label & CR Example (rationale in bold) \\
\midrule
\multirow{10}{*}{-} & \nl{i 've tried the belkin fm transmitter unit with it \& it worked well when i set it on top of a portable radio , but was awful trying to use \textbf{in the car} which is somewhat of a disappointment .} \\
& \nl{but the major problem i had was with the \textbf{software} .} \\
& \nl{after a week i tried to load some more songs and delete a few but the \textbf{auto load} didn 't do anything but turn on my player .} \\
& \nl{2 . the scroll button is n 't the best , as it sometimes can be \textbf{hard to select} .} \\
& \nl{iriver has a better fm receiver built in , but the drawback to iriver products is they are \textbf{flimsy and poorly constructed} .} \\
& \nl{i would imagine this is a problem with any camera of a \textbf{compact nature} .} \\
& \nl{the \textbf{pictures are a little dark} sometimes .} \\
& \nl{the \textbf{depth adjustment} was sloppy .} \\
& \nl{the \textbf{instructions} that come with it do n 't explain how to make things simple .} \\
& \nl{my " fast forward " button works , but it \textbf{takes a little extra pressure} on it to make it go .} \\
\midrule
\multirow{10}{*}{+} & \nl{i did not conduct a rigorous test , but just took some identical shots in \textbf{identical lighting} with both cameras , and the canon won hands down .} \\
& \nl{as a whole , the dvd player has a \textbf{sleek design and works fine} .} \\
& \nl{i , as many others , have waited for many years for the \textbf{convergence of price} , features , size and ease of use to hit that happy center point .} \\
& \nl{+ i had \textbf{no problem} using musicmatch software already on my computer to load songs and albums onto this unit} \\
& \nl{apex is the best \textbf{cheap quality} brand for dvd players .} \\
& \nl{i chose this one because from what i read , it was the \textbf{best deal for the money} .} \\
& \nl{the \textbf{two-times optical zoom} operates smoothly and quietly , and lo and behold , a two-piece shutter-like cap automatically slides closed over the lens when you turn the camera off .} \\
& \nl{this camera is perfect for the person who wants a compact camera that \textbf{produces excellent photos} in just about any situation .} \\
& \nl{it was easy enough to remove the front plate , and there was only one way the \textbf{battery could be inserted} .} \\
& \nl{i have been very impressed with my purchase of the sd500 i bought it at the beginning of the month as \textbf{the ultimate pocket camera} and have shot 300 images so far with it .} \\
\bottomrule
\end{tabular}
\caption{Rationales (in bold) extracted by the zero-shot \quip{} sentiment analysis model for the Customer Reviews (CR) dataset.
We show ten random examples for each label on which the model made the correct prediction.}
\label{tab:cr-rationales} 
\end{table*}